\documentclass[letterpaper, 10 pt, conference]{ieeeconf}  

\IEEEoverridecommandlockouts                              

\usepackage{graphics,graphicx}           
\usepackage{amsmath,amssymb}    
\usepackage{color,xspace}
\usepackage{subfigure}
\usepackage{balance}
\usepackage{cite}
\usepackage{color,soul}
\usepackage{mathtools}
\usepackage{makeidx}
\usepackage{robustindex}
\usepackage{float}
\usepackage{multirow}
\usepackage{textcomp}
\usepackage{bm}
\usepackage{trimclip}

\makeindex

\title{\LARGE \bf
Flappy Hummingbird: An Open Source Dynamic Simulation of Flapping Wing Robots and Animals
}

\author{Fan Fei, Zhan Tu, Yilun Yang, Jian Zhang, and Xinyan Deng
\thanks{The authors are with the School of Mechanical Engineering, Purdue University. (Email: xdeng@purdue.edu).}
\thanks{The code is available at (https://github.com/purdue-biorobotics/flappy).}
}

\begin{document}

\maketitle
\thispagestyle{empty}
\pagestyle{empty}

\begin{abstract}

Insects and hummingbirds exhibit extraordinary flight capabilities and can simultaneously master seemingly conflicting goals: stable hovering and aggressive maneuvering, unmatched by small scale man-made vehicles. Flapping Wing Micro Air Vehicles (FWMAVs) hold great promise for closing this performance gap. However, design and control of such systems remain challenging due to various constraints. Here, we present an open source high fidelity dynamic simulation for FWMAVs to serve as a testbed for the design, optimization and flight control of FWMAVs. For simulation validation, we recreated the hummingbird-scale robot developed in our lab in the simulation. System identification was performed to obtain the model parameters. The force generation, open-loop and closed-loop dynamic response between simulated and experimental flights were compared and validated. The unsteady aerodynamics and the highly nonlinear flight dynamics present challenging control problems for conventional and learning control algorithms such as Reinforcement Learning. The interface of the simulation is fully compatible with OpenAI Gym environment. As a benchmark study, we present a linear controller for hovering stabilization and a Deep Reinforcement Learning control policy for goal-directed maneuvering. Finally, we demonstrate direct simulation-to-real transfer of both control policies onto the physical robot, further demonstrating the fidelity of the simulation.

\end{abstract}

\section{Introduction}

\begin{figure}[!b]
\vspace{-0.2in}
\centering
\includegraphics[trim = 0mm 0mm 0mm 0mm,width=0.9\columnwidth]{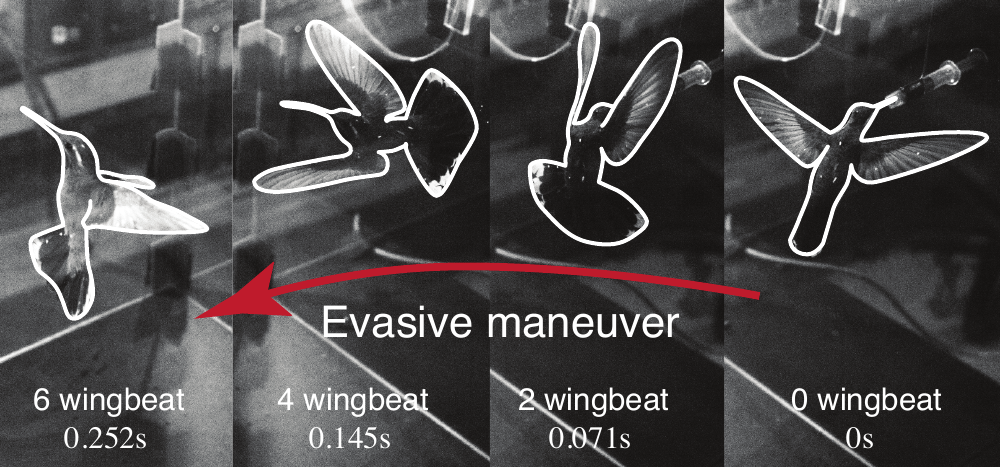}
\caption{A hummingbird flying from stable hovering to maneuvering back to hovering under 8 wingbeats \cite{cheng2016flight}. The silhouette of the hummingbird is enhanced for better visibility.}
\label{fig:hummingbird}
\end{figure}

Flying animals possess extraordinary capabilities and demonstrate rich repertoire of agile maneuvers, often under a variety of disturbances such as wind gust and rain \cite{ellington1984aerodynamics}. They remain surprisingly stable during hover and can make sharp turns in a split second, e.g. the escape maneuvers of hummingbird take only 8 wing beats - a quarter of a second - to complete, as shown in Fig. \ref{fig:hummingbird}. This is unmatched by man-made counterparts. Great progress has been made in recent years in the development of Flapping Wing Micro Air Vehicles (FWMAVs), among which Delfly \cite{de2009design}, RoboBee \cite{ma2013controlled}, Nanohummingbird \cite{keennon2012development}, KUBeetle \cite{phan2017design}, COLIBRI \cite{roshanbin2017colibri} and Purdue Hummingbird robot\cite{zhang2017design} have demonstrated successful takeoff and hovering.

Due to the complex nature of the unsteady aerodynamics during high-frequency flapping motion, the development of such platforms to match the performance of the nature's flyers remains extremely challenging. On the design side, the system design and optimization problems are further complicated by the stringent weight, size and power constraints \cite{zhang2017design}. On the control side, the unsteady aerodynamics, high-frequency flapping oscillations, and noisy nonlinear flight dynamics present some extreme hurdles for maneuver control \cite{zhang2017geometric}. In summary, substantial progress is needed in all aspects of the system before a truly bio-inspired vehicle can be developed to approach the performance of its biological counterpart.

Furthermore, the difficulties and the limited availability of such hardware platforms could deter or slow down the interest and progress in FWMAVs. As a comparison, conventional robotic platforms are much more accessible such as manipulators, ground vehicles, underwater robots, legged robots, and drones/quadcopters. There are also various simulation and analytical tools for interested researchers to test their ideas \cite{quigley2009ros, todorov2012mujoco, brockman2016openai, koenig2004design}. However, there is yet to be an easy-to-use flapping wing MAV simulation toolkit.



\begin{figure}[!t]
\centering
\includegraphics[trim = 0mm 0mm 0mm 0mm,width=0.99\columnwidth]{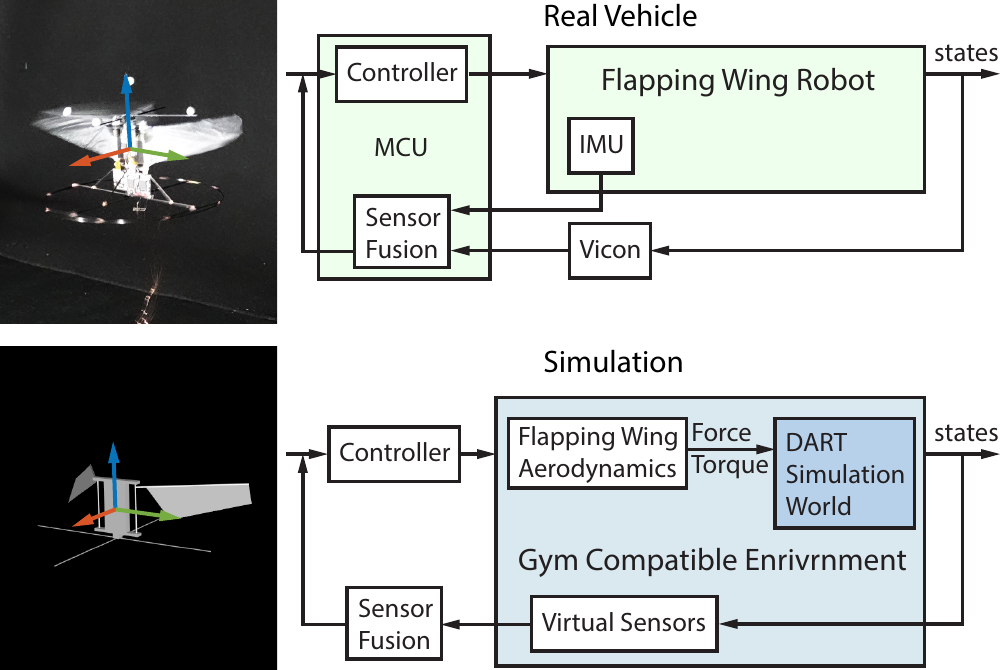}
\caption{Diagrams of the FWMAV robot platform and its simulation environment.}
\label{fig:sys}
\vspace{-0.2in}
\end{figure}

To facilitate the design of FWMAV platforms and the study of flapping flight control in general, we present an open source high fidelity dynamic simulation for FWMAVs and flapping-wing animals such as hummingbirds and insects. Using the flapping-wing robot developed in our lab \cite{zhang2017design} as a blueprint, we built its virtual counterpart in the simulation environment. The simulation is written in C++ with Python binding, using customized flapping-wing aerodynamic models and DART \cite{lee2018dart} physics engine to solve multi-body kinematics and dynamics. The physical parameters were obtained by performing system identification on the robot. The aerodynamic modeling is validated through wing kinematics and force/torque measurements. Open loop flight tests were conducted and state transition statistics was verified. Finally, we demonstrate that the fidelity of the simulation is suitable for continuous control tasks. A feedback flight controller is designed in the simulation to achieve stable position tracking for the robot. When transfer to the robotic platform, the same flight performance was achieved on the vehicle by directly implementing the simulated controller onboard the robot. We also developed a goal-directed flight maneuvering control policy using deep reinforcement learning. The policy was optimized in simulation and directly transferred to the robot. Successful transferring of both controllers further validates the fidelity and effectiveness of the simulation.

With this tool, MAV designers can iterate and optimize their design and parameters before tediously building physical variations and testbeds, as the vehicle dynamics is detailed down to component level in simulation. This tool is built on top of DART, so topics like state estimations, perception, localization and mapping, can be studied with integration to ROS and Gazebo. We also challenge control and learning researchers to control such system to be equal or better at maneuvering than the animal, for which, we provide hummingbird data for comparison (Fig. \ref{fig:hummingbird}). Faster than realtime simulation and OpenAI Gym interface support research topics on Control Theory, Deep Reinforcement Learning and Imitation Learning. For experimental biologists, we provide several flapping-wing animal models with full degrees of freedoms of the wing motion, aiding the study of neural muscular control, flapping flight behaviors and evolution. We are open to provide experimental support on the physical robots for simulation users, like Robotarium \cite{pickem2017robotarium} and DuckieTown \cite{paull2017duckietown}. The code and data will be available online.

\section{System Model}

\subsection{System Definition}

The robotic vehicle used in this study is a motor-driven FWMAV, on which two motors were equipped to drive the two wings independently. It has a wingspan of $168mm$ and weights $12g$. Torsion spring was used to achieve resonance. Details of the platform are presented in \cite{zhang2017design}. We use wingbeat modulation technique to generate thrust and control torque\cite{doman2010wingbeat}. The four input signal is defined as amplitude $V_{amp}$, amplitude difference $V_d$, bias $V_0$ and split-cycle parameter $\delta\sigma$, which controls thrust, roll, pitch, and yaw torque.

To recreate the FWMAV platform in the simulation, we describe the vehicle with five rigid bodies: one torso, two leading edge frames, and two wings. The leading edge is linked to the torso with a stroke joint, and the wing is linked to the leading edge with a rotation joint. The stroke joints are configured with spring constants. We simulate motor torques to drive the leading edge frame back and forth. Aerodynamic forces and torques can be calculated and applied on the wings at the span-wise and chord-wise center of pressure $r_{cp}$ and $d_{cp}$. The stroke and rotation angles are set with limited movement range.
\begin{figure}[!t]
\vspace{0.02in}
\centering
\includegraphics[trim = 0mm 0mm 0mm 0mm,width=0.95\columnwidth]{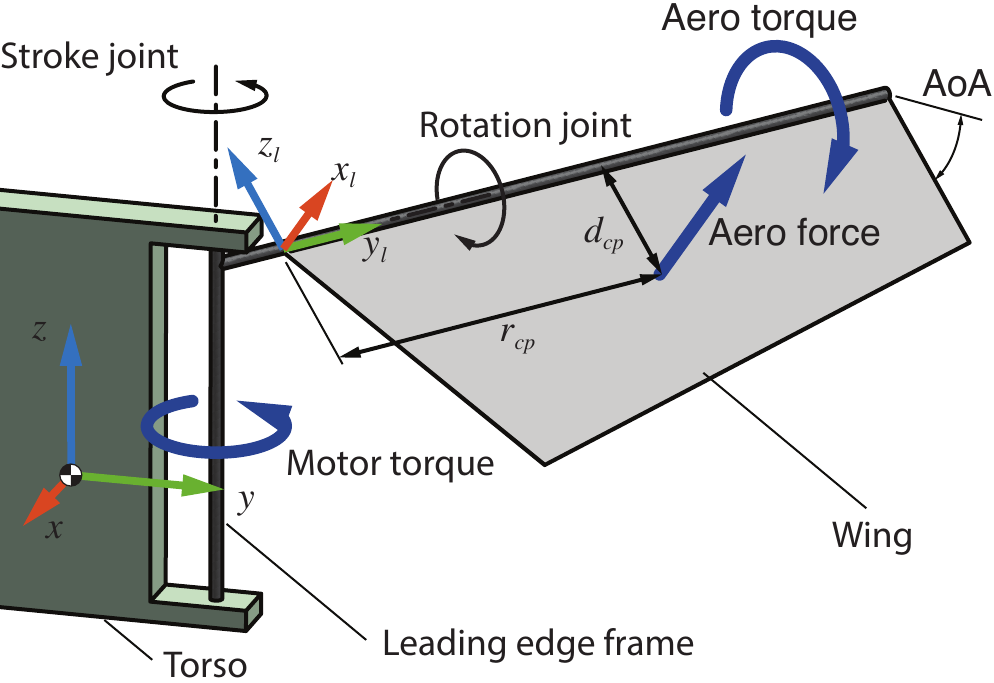}
\caption{Coordinate frames of the flapping wing in simulation, with the applied forces and torques illustrated. The angle of attack (AoA) formed by passive rotation of the wing, the torso, the left leading edge and the left wing joint are shown.}
\label{fig:joint_link}
\vspace{-0.2in}
\end{figure}

\begin{figure}[!h]
\centering
\includegraphics[trim = 0mm 0mm 0mm 0mm,width=1.0\columnwidth]{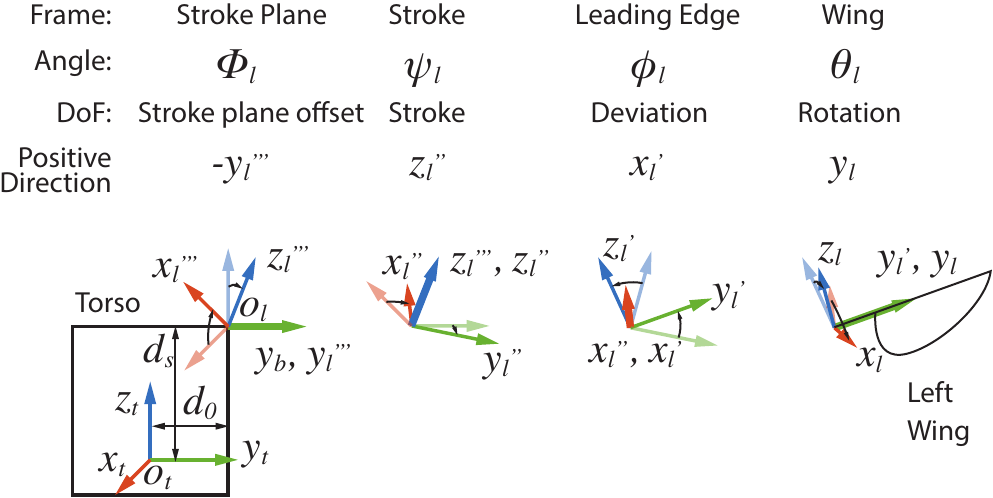}
\caption{Coordinate frame definition of the left wing's movement. The origins of all wing frames are located at shoulders $o_l$. The four degrees of freedom is used to describe the wing kinematics of both robots and animals.}
\label{fig:coord}
\vspace{-0.2in}
\end{figure}

We define the wing movement with four degrees of freedoms so the aerodynamic model can be generalized to robots and animals. The leading edge has three degrees of freedoms: stroke plane offset $\Phi$, stroke angle $\psi$, deviation angle $\phi$, and the wing has one degree of freedom $\theta$, which is the rotation angle. As shown in Fig. \ref{fig:joint_link}, $\Phi$ and $\phi$ are fixed at zero for our vehicle platform.
The coordinate system of the body and the left wing is illustrated in Fig. \ref{fig:coord}, where $o_t$ is the center of mass (CoM), $o_l$ is the left shoulder, $d_s$ is the distance from CoM to the shoulder and $d_0$ is half shoulder width. The positive direction of each degree of freedom is defined such that for both wings, positive $\Phi$ can produce positive yaw torque, positive $\psi$ corresponds to upstroke, positive $\phi$ corresponds to heaving or abduction and positive $\theta$ corresponds to pronation.


\subsection{Aerodynamics}

To accurately capture the body dynamics of the vehicle, we need to calculate the instantaneous aerodynamic forces and torques. Based on the blade element method and quasi-steady model, we calculate the normal force and rotational moment on the wing from the effective wing kinematics by incorporating body kinematics into wing motion through coordinate transformation.
To calculate the aerodynamic force we need the velocity of the wing at span-wise location $r_w$, and its angle of attack $\alpha$. For convenience, we divide the velocity into two components: one out-of-plane component that is normal to the $x'-y'$ plane, one in-plane component within the $x'-y'$ plane and normal to the leading edge.
\begin{equation}\label{eqn:wing_vel}
\begin{aligned}
	u_i^k &= u_{i1}^kr_w + u_{i0}^k\\
	u_o^k &= u_{o1}^kr_w + u_{o0}^k
\end{aligned}
\end{equation}
where $k=0$ indicates left wing and $k=1$ indicates right wing, subscript $i$ indicates in-plane component, $o$ indicates out of plane component.

For the left wing, coefficient $u_{i1}^L = \omega_{z'}$, $u_{i0}^L = v_{x}$, $u_{o1}^L = \omega_{x'}$ and $u_{o0}^L = -v_{z'}$. $u_i^L$ is the in-plane component, where its sign follows the positive stroke direction (right hand about $z'$), and $u_i^L$ is the out-of-plane component, defined along the $z'$ axis; $\mathbf{v}_{oL}' = [v_{x'}, v_{y'}, v_{z'}]^T$ and $\bm{\omega}_L' = [\omega_{x'}, \omega_{y'}, \omega_{z'}]^T$ are the linear and angular velocity of the left leading edge in $x'y'x'$ frame.

The body linear and angular velocity is $\mathbf{v}_{ob} = u\hat{i} + v\hat{j} + w\hat{k}$ and $\bm{\omega}_{ob} = p\hat{i} + q\hat{j} + r\hat{k}$. The left shoulder velocity in the body frame can be calculated by
\begin{equation}
	\mathbf{v}_{oL}^b = \mathbf{v}_{ob} + \bm{\omega}_{ob} \times \mathbf{r}_{oL/ob}
\end{equation}

To get the velocity components in the leading edge $x'y'z'$ frame, we define a rotation matrix$\begin{bmatrix}
		R_L^T
	\end{bmatrix}$ which first rotates about $-y'''/-y_{bL}$ axis then $z'''/z''$, then finally $x''/x'$ axis.
The left shoulder velocity in the leading edge frame is
\begin{equation}\label{eqn:wing_lin_vel}
	\mathbf{v}_{oL}' = \begin{bmatrix}
		R_L^T
	\end{bmatrix}\mathbf{v}_{oL}^b
\end{equation}


The angular velocity of the leading edge frame is
\begin{equation}
	\bm{\omega}' = \bm{\omega}_{ob} - \dot{\Phi}_L\hat{j}_{bL} + \dot{\psi}_Lk''' + \dot{\phi}_L\hat{i}''
\end{equation}
Knowing $\hat{i}'' = \hat{i}'$ and $\hat{k}''' = \hat{k}'' = \hat{k}'C\phi_L + \hat{j} S\phi_L$, $\bm{\omega}'$ can be expressed in $x'y'z'$ frame as
\begin{equation}\label{eqn:wing_ang_vel}
	\bm{\omega}' = \begin{bmatrix}
		R_L^T
	\end{bmatrix} \begin{bmatrix}
		p\\
		q-\dot{\Phi}_L\\
		r
	\end{bmatrix} + \begin{bmatrix}
		\dot{\phi}_L\\
		\dot{\psi}_L\sin\phi_L\\
		\dot{\psi}_L\cos\phi_L
	\end{bmatrix}
\end{equation}
With (\ref{eqn:wing_lin_vel}) and (\ref{eqn:wing_ang_vel}), the coefficient $u_1$ and $u_0$ in equation (\ref{eqn:wing_vel}) can be calculated. Right wing velocity is calculated similarly.

For simplicity and computation efficiency, we consider the angle of attack at the span-wise center of pressure $r_{cp} = R_w\hat{r}^3_3/\hat{r}^2_2$
\begin{equation}
	\alpha = \theta + \text{sgn}(u_i)\frac\pi 2 - \text{atan}\left(\frac{u_{o1}r_{cp} + u_{o0}}{u_{i1}r_{cp} + u_{i0}}\right)
\end{equation}
Define normal force $F_N$ in $x$ direction, aerodynamic moment $M_{aero}$ and rotational damping moment $M_{rd}$ in right-hand $y$ direction, from observation, we have $\text{sgn}(\alpha) = \text{sgn}(F_N) = -\text{sgn}(M_{rd})$.

To calculate the forces, the velocity squared at $r_w$ can be written as
\begin{equation}
\begin{aligned}
	u^2 &= u_i^2 + u_o^2\\
	&=a_{u2}r_w^2 + a_{u1}r_w + a_{u0}
\end{aligned}
\end{equation}
Integrate blade element force along the wingspan \cite{whitney2010aeromechanics}, the normal force, aerodynamic moment and rotational damping moment are
\begin{equation}
\begin{aligned}
	F_N =& \frac12\rho_A C_N(\alpha)\bar{c}\left(a_{u2}R_w^3\hat{r}_2^2 + a_{u1}R_w^2\hat{r}_1^1 + a_{u0}R_w\hat{r}_0^0\right)\\
	M_{aero} =& -\frac12\rho_A \hat{d}_{cp}(\alpha)C_N(\alpha)\bar{c}^2\big(a_{u2}R_w^3\hat{z}_{cp}^2\\
	&+ a_{u1}R_w^2\hat{z}_{cp}^1 + a_{u0}R_w\hat{z}_{cp}^0\big)\\
	M_{rd} =& -\frac18\rho_A|\dot{\theta}|\dot{\theta}C_{rd}R_w\bar{c}^4\hat{z}_{rd}
\end{aligned}
\end{equation}
where a non-dimensional chord-wise center of pressure $\hat{d}_{cp}$ is adopted from \cite{dickson2006integrative} and implemented as $2\pi$ periodic.




The total instantaneous aerodynamic forces applied on the wing are $M_{rd}$ about $y$ axis and $F_N$ at $d_{cp} = -M_{aero}/F_N$ cord-wise and $r_{cp}$ span-wise.

\section{Model Validation}

It is well known that flapping wing robots are sensitive to mechanical imperfections in force production \cite{chirarattananon2014adaptive, zhang2017geometric}. To verify the fidelity and accuracy of the simulation quantitatively, it is ideal to have a model identical to the real robot. We first conduct system identification to tweak the uncertain system parameters to best approximate the mechanical trim condition of the real robot, then we validate its wing kinematics and open loop state transition. Note that for controller and vehicle design, small parametric uncertainty is acceptable, as the overall dynamic behavior is not affected and the small mechanical trim can be compensated by the controller. For reinforcement learning applications, dynamics randomization can be used to achieve a robust control policy \cite{peng2018sim, tobin2017domain}.

%
%
%
%

\subsection{System Identification and Force Mapping}
Most system parameters can be directly measured and stay constant such as motor torque constant as well as mass and wing shape parameters if assume no physical damage occurs. Some parameters cannot be measured accurately but have non-negligible effects on body torque generation, such as spring resting position and wing rotation angle limits, which create net pitch and yaw torques. We measured all parameters, and use system ID to tweak the uncertain parameters within small bounds.
Given the large number of uncertain parameters and highly coupled nonlinear dynamics, we use genetic algorithm to find the best fit. The parameters to be adjusted are: motor resistance ${R_m}_l$ and ${R_m}_r$, spring stiffness in both directions ${K_s}_l^+$, ${K_s}_l^-$ and ${K_s}_r^+$, ${K_s}_r^-$, mid-stroke resting angle ${\psi_0'}_l$ and ${\psi_0'}_r$ and wing rotation angle upper and lower limit ${\Theta^+}_l$, ${\Theta^-}_l$, ${\Theta^+}_r$ and ${\Theta^-}_r$.

Since the mass property of the robot is largely constant and can be easily measured, the system identification process focuses more on accurate force generation.
We use an ATI Nano 17 sensor to measure the cycle-averaged force generated by the robot under different operating points. A total of 37 different inputs were used and 6 body force and torque were measured at each operating point. This force map with 222 data points will be used as the ground truth to measure the accurateness of the force generation of the simulation. The cost is defined as the squared error sum between the measured force and the force calculation from simulation across all data points.
\begin{figure}[!tb]
\centering
\includegraphics[trim = 0mm 0mm 0mm 0mm,width=0.9\columnwidth]{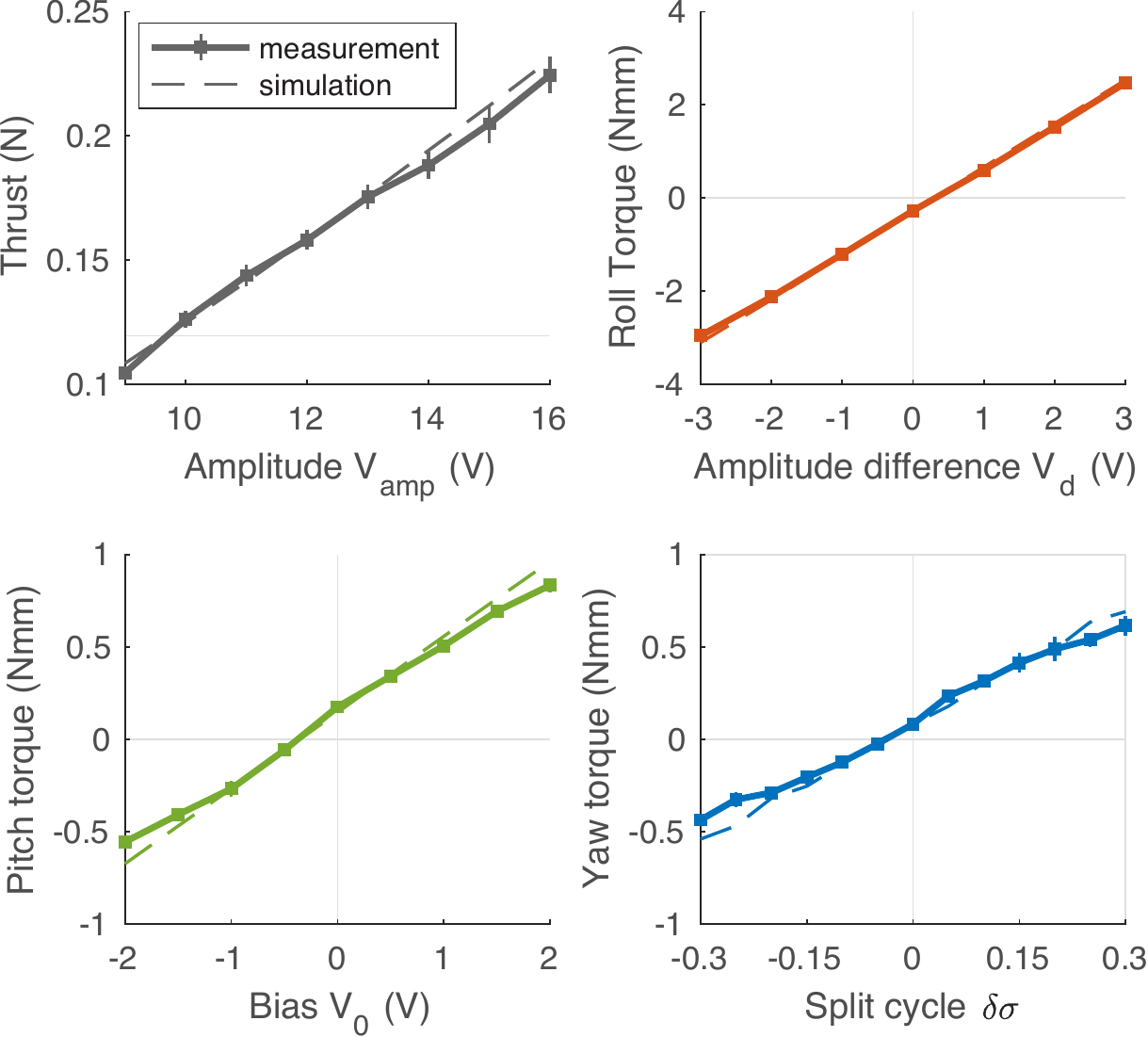}
\caption{The force map of the real and the simulated vehicle. Thrust and three control torques are matching well after system identification.}
\label{fig:force_map}
\vspace{-0.2in}
\end{figure}
The parameters were optimized with 200 individuals for 200 generations. The result with the best fit is shown in Fig. \ref{fig:force_map}. The simulated force map matches the measurement well, with minor errors under larger inputs. This could be caused by the nonlinearity of the spring at a large deviation angle. The total error is 4.1\%.

\subsection{Wing Kinematic}
To further validate the system identification, we compare the wing kinematics of the real vehicle with the simulation. A high-speed camera is used to record the wing motion at $5000fps$, wing stroke and rotation angles are extracted using \cite{hedrick2018digitize}. The real wings have a bi-stable design with the majority of the area constructed as a rigid plate. Since they still have a certain degree of twist, we pick the wing tip for rotation angle measurement since tip velocity is the highest. The simulated robot has a constant geometric AoA.

\begin{figure}[!tb]
\centering
\includegraphics[trim = 0mm 0mm 0mm 0mm,width=0.9\columnwidth]{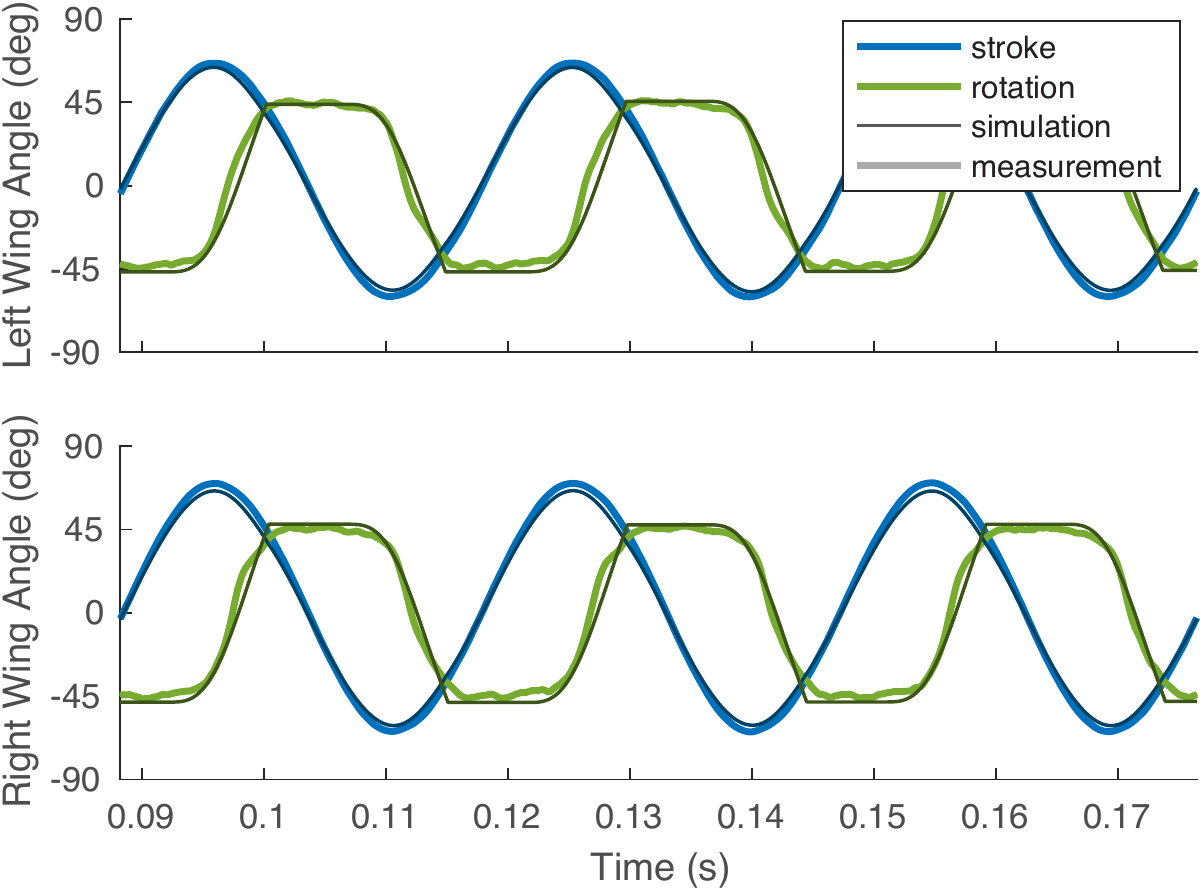}
\caption{A sample wing kinematics comparison of left and right wings between the robot and simulation with $10V$ input.}
\label{fig:wing_kin}
\vspace{-0.1in}
\end{figure}

The wing kinematics under sinusoidal input is shown in Fig. \ref{fig:wing_kin}. As seen from the figure, larger stroke amplitude on right wing corresponds to negative roll torque in force measurement, positive bias in stroke angle correlates to the positive pitch trim, and the difference in the rotation angle limit between two wings will result in a net yaw torque.

%
%

\subsection{Open Loop State Transition}
\begin{figure}[!tb]
\begin{minipage}{0.95\columnwidth}	
\includegraphics[trim = 0mm 0mm 0mm 0mm,width=1\columnwidth]{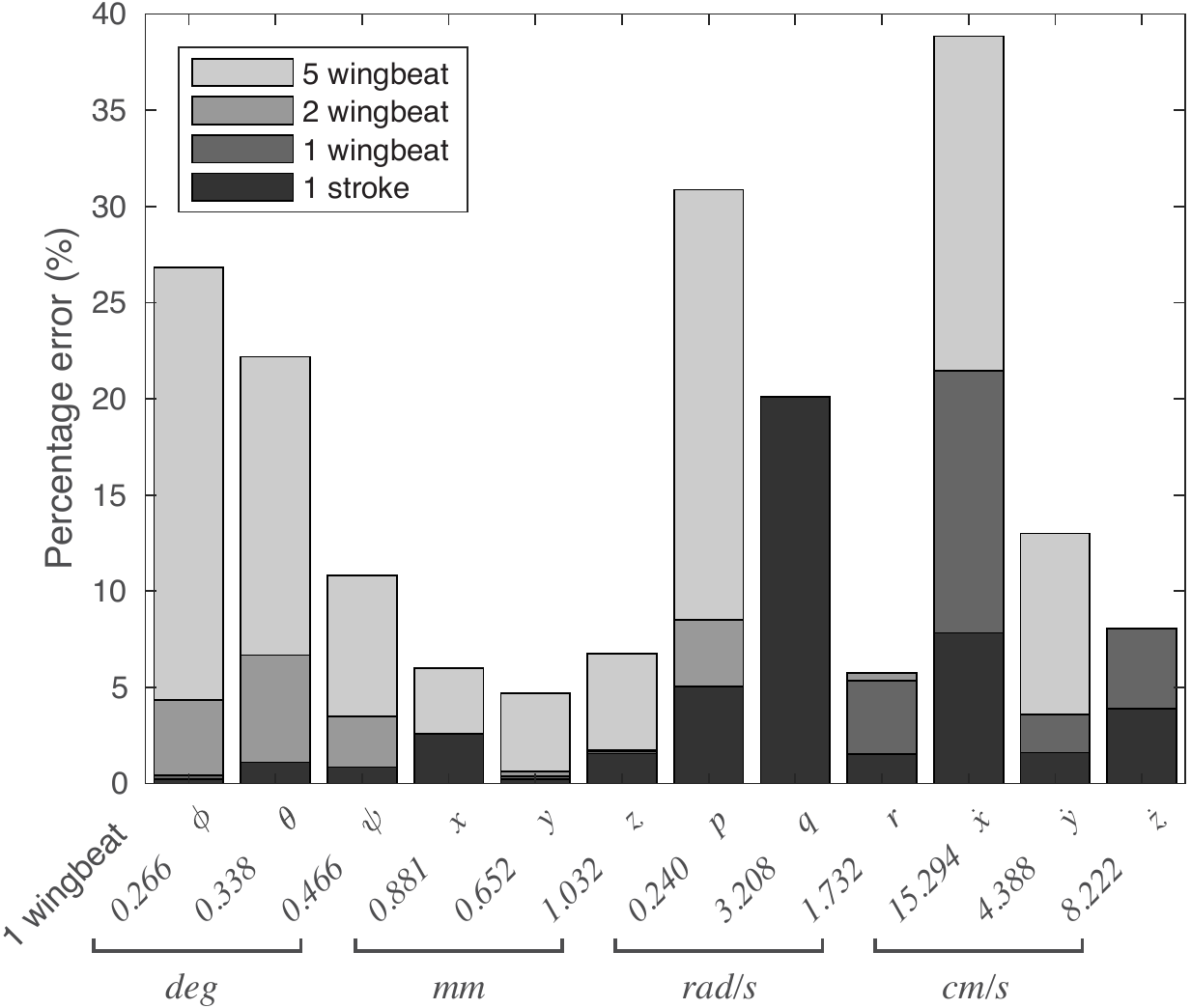}
\end{minipage}
\caption{The normalized state transition error between simulation and real robot is shown. The averaged absolute error after 1 wingbeat of each state variable is listed at the bottom.}
\label{fig:ol_stat}
\vspace{-0.2in}
\end{figure}
For continuous control, the behavior of a vehicle can be viewed as a Markov decision process (MDP) with state space $\mathcal{S} \subseteq \mathbb{R}^{12}$, action space $\mathcal{A} \subseteq \mathbb{R}^{2}$. For the simulation to be statistically meaningful, we need to evaluate whether the open loop state transition dynamics $p_o(s_{t+1}|s_t,a_t)$ of the simulation matches that of the real vehicle. The state of the vehicle is $s_t=[\phi, \theta, \psi, x,y,z,p,q,r,\dot{x},\dot{y},\dot{z}]^T$ and the action is $a_t=[V_l,V_r]^T$. A total of 20 open loop flights were conducted, to avoid ground effect, only data points with altitude of at least five wing chord length were used. A total of 2500 valid samples were collected.

To evaluate the simulation state transition, we use each sample from the flight data as the initial state, and run the simulation with the recorded input and compare the state values with measurements after a given time. 
The averaged result is shown in Fig. \ref{fig:ol_stat} for each state. The error is normalized by the maximum range within one wingbeat across the 2500 samples collected for each state. The error shows that the simulation can accurately capture the state transition within one wingbeat with less than 5\% error, where Euler angle error is smaller than $1^\circ$ and position error about $1mm$. The state transition error is still acceptable after 2 wingbeats with only pitch and $x$ velocity showing larger error. This is expected as pitch and $x$ direction corresponds to the severe body vibration caused by the cyclic aerodynamic forces.

\section{Flight Control Baselines with Experimental Validation}
\subsection{Closed Loop Position Controller}
\begin{figure*}[!ht]
\centering
\includegraphics[trim = 0mm 0mm 0mm 0mm,width=1\textwidth]{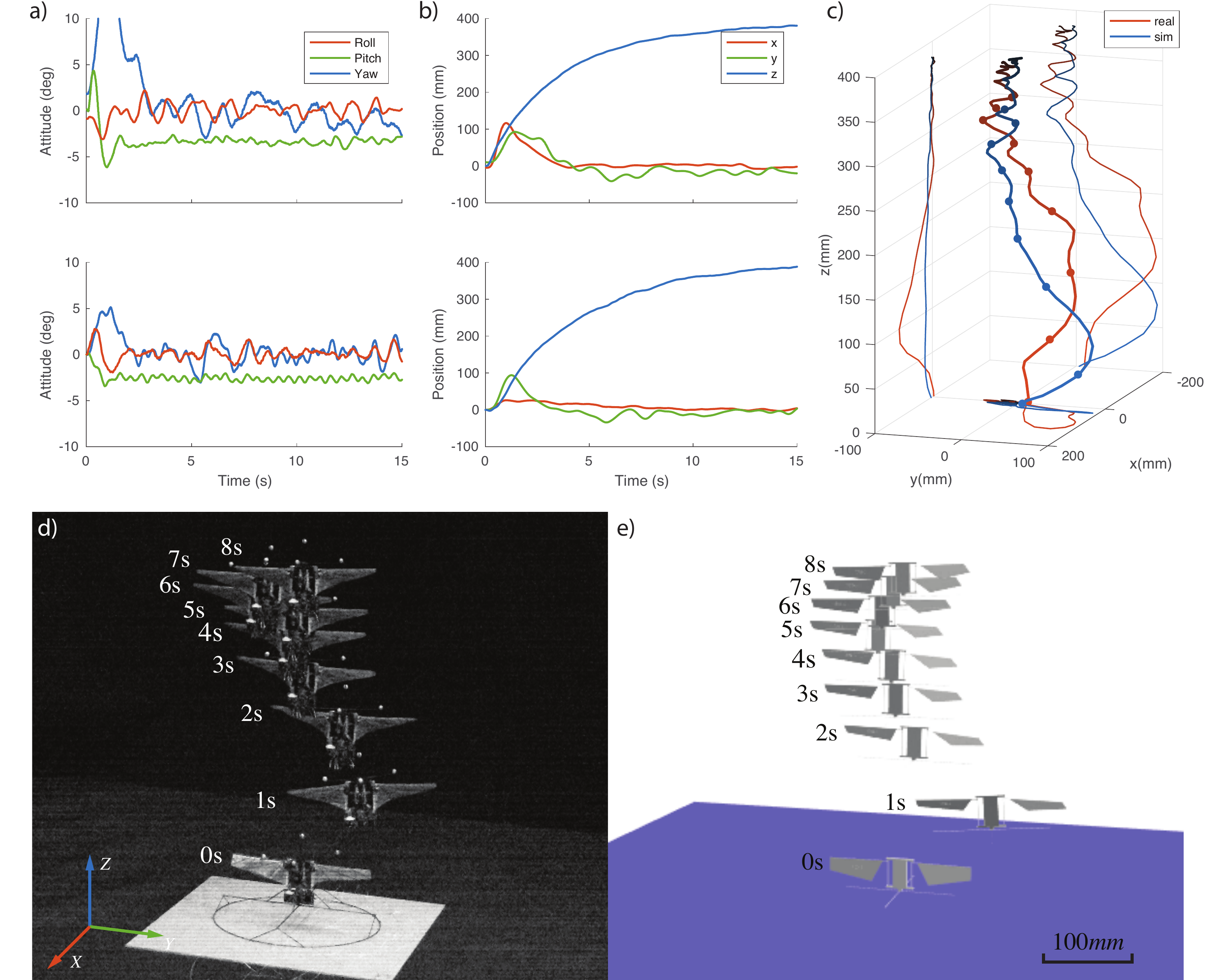}
 \caption{a,b) These two figures demonstrate the body Euler angles and the position of real (top) and simulated (bottom) robots, respectively. c) Plots of the positions of both vehicles in the inertia frame; the dots indicate the first 8 seconds of the flight. d,e) Composed image sequences of the closed loop controlled flights of the real (left) and simulated (right) robots. The first 8 seconds of the flight was shown, demonstrating direct sim-to-real transfer of the controller and control gains.}
\label{fig:cl}
\vspace{-0.2in}
\end{figure*}
\begin{figure*}[!ht]
\centering
\includegraphics[trim = 0mm 0mm 0mm 0mm,width=0.95\textwidth]{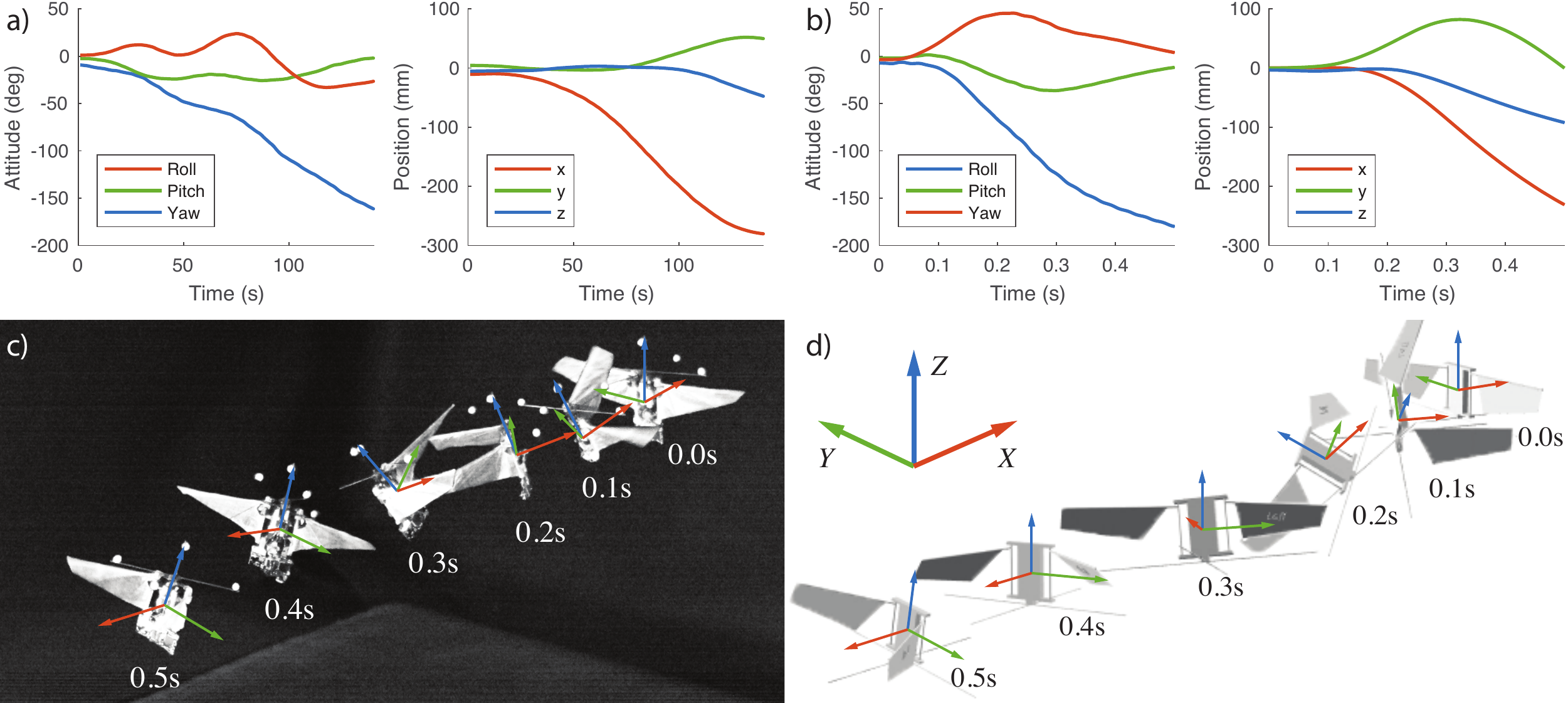}
 \caption{a,b) These two figures demonstrate the body Euler angles and the position of real (left) and simulated (right) robots, respectively. c,d) Composed image sequences of the controlled flights of the real (left) and simulated (right) robots. Direct sim-to-real transfer of the reinforcement learning maneuver policy is shown here.}
\label{fig:maneuver}
\vspace{-0.1in}
\end{figure*}

We now show that the simulation can be used for flight controller design. We constructed a simple PID flight controller based on rigid body dynamics for the FWMAV in the simulation. The controller has a cascading structure wherein the outer loop is a cascading position and velocity PD controller that generates the target attitude, and the inner loop is the PID attitude controller. Heading (yaw) was controlled independently. The simulation ran at 10kHz, virtual Vicon and IMU sensors were implemented in the simulation at 150Hz and 500Hz respectively, with noise characteristics and delay similar to their physical counterpart. The sensor fusion and control algorithm were run at 500Hz. We manually tuned the control gains to achieve a stable flight. The controller with the tuned gains was then transferred to the robotic FWMAV platform.

The closed loop performance of the robotic vehicle is similar to that observed in the simulation as expected. Closed loop control error is very close between the two as shown in Table \ref{tab:rms}. A sample flight data from the vehicle and the simulation using the same controller with the same reference input is shown in Fig. \ref{fig:cl}. The vehicle is commanded to takeoff and hovering at the height of 0.4m. Observing Fig. \ref{fig:cl} a) and b), each axis exhibits similar behavior and tracking error, indicating the closed loop dynamics is very similar between simulation and experiment. Both vehicles move to their left during takeoff, as a result of the negative roll torque offset. The non-zero pitch angle is due to a small thrust component in the $x$ direction. These phenomenons further justifying the accuracy of the simulation and system identification.
\begin{table}[!t]
\vspace{-0.0in}
\caption{Closed loop control error}
\begin{center}
\begin{tabular}{ c | c c c c c c}
RMS (deg\&mm) & Roll & Pitch & Yaw & $x$ & $y$ & $z$\\
\hline
Experimental & 1.76 & 3.95 & 5.23 & 34.9 & 37.2 & 24.4\\
 Simulation & 1.60 & 3.86 & 4.79 & 36.5 & 38.7 & 26.9
\end{tabular}
\end{center}
\label{tab:rms}
\vspace{-0.2in}
\end{table}

\subsection{Goal-directed Maneuvering}

To further demonstrate the fidelity of the simulation, we present a reinforcement learning policy transfer for maneuvering flight of the FWMAV. The goal of the flight maneuver is to move from position $\mathbf{p}_0 = [0,0,0]^T(m)$ with yaw heading $\psi_0 = 0^\circ$ to $\mathbf{p}_f = [-0.21,0,0]^T (m)$ with yaw heading $\psi_f = 180^\circ$, in hope of mimicking the hummingbird's fast escape maneuver \cite{cheng2016flight}.

We use a standard reinforcement learning setup to optimize a maneuvering policy approximated by a standard MLP. The state transition dynamics $p_c(s_{t+1}|s_t,a_t)$ is the closed-loop dynamics of the vehicle with feedback controller. The input is the state $s_t$, and the output in this case is the additional control effort $a_t = [\Delta V_{amp}, \Delta \delta V, \Delta V_0, \Delta \delta\sigma]^T$. The reward is selected such that the vehicle will receive positive reward near $\mathbf{p}_f$ and with correct heading.

\begin{figure}[!tb]
\centering
\includegraphics[trim = 0mm 0mm 0mm 0mm,width=0.9\columnwidth]{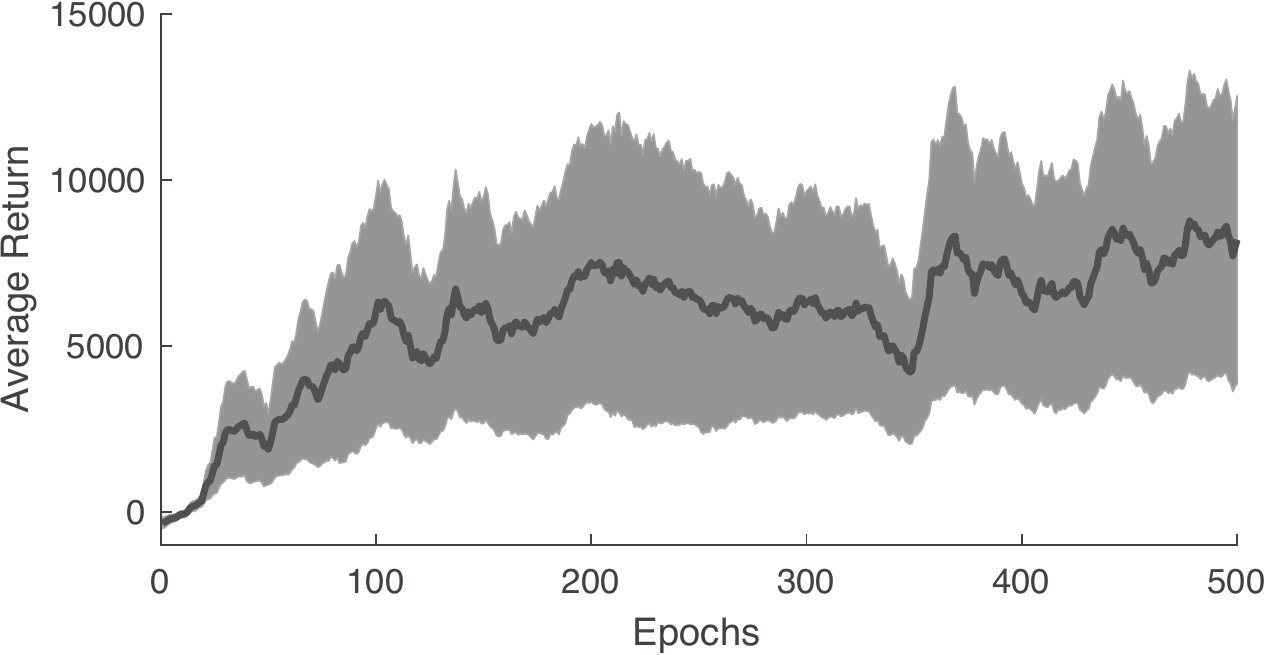}
\caption{Training curve of the maneuver policy averaged over 5 random seeds.}
\label{fig:learning_curve}
\vspace{-0.1in}
\end{figure}

Since the system is largely deterministic, popular actor-critic algorithm deep deterministic policy gradient (DDPG) is selected to train the policy. We use 2 hidden layers of 32 hidden units for the actor network and 2 hidden layers of 64 hidden units for the critic network. The implementation is based on \cite{duan2016benchmarking} with same hyperparameters as \cite{lillicrap2015continuous}. Dynamics randomization \cite{peng2018sim} is used during training, where we randomize the physical parameters of the vehicle slightly to improve the robustness of training and ensure simulation to the real world transfer. The learning curve averaged over 5 runs with different random seeds is shown in Fig. \ref{fig:learning_curve}.

The trajectory manifested by the optimized policy is a unique multi-axis fast maneuver that minimizes the travel time. The resulting flight from simulation and the real flight is shown in Fig. \ref{fig:maneuver}. Detail of this study is presented in \cite{fei2019learning}.

\section{CONCLUSIONS}

In this work, we developed an open source high fidelity simulation with realistic multi-body dynamics for flapping wing flight. Instantaneous aerodynamics were simulated using blade element theory and quasi-steady aerodynamic model which was validated by force measurements. Open loop state transition dynamics of the simulation is validated by calculating the state transition error between the simulation and the vehicle. The error shows the simulation can accurately predict instantaneous state transitions and capture the dynamic effects. With successful sim-to-real transferring, we demonstrate the fidelity of the simulation in two controller design applications: 1) a linear cascading PID controller for FWMAV position control, 2) unique goal-directed maneuvering of FWMAV using a policy optimized by reinforcement learning. For both applications, no special treatments were needed in controller implementation. The experimental data match simulation results, proving the fidelity of the simulation. With motor and contact dynamics, the current feedback could also be used as tactile sensing to mimic animal somatosensory \cite{tu2019acting}, which could be exploited in simulation for control and trajectory planning design. This open source simulation can serve as a design and flight control testbed for scientists and researchers interested in studying flapping wing animals and robots. The code, baselines, and data will be available online, and experimental support on the robot will be provided.







\balance

\bibliography{all}
\bibliographystyle{IEEEtran}

\end{document}